\pdfoutput=1
\documentclass{article} % For LaTeX2e
\usepackage{times}
\usepackage{verbatim}
\usepackage{graphicx}
\usepackage{subfig}
\usepackage{amsmath}
\usepackage{amsthm}
\usepackage{amsfonts}
\usepackage{amssymb}
\usepackage{color}
\usepackage{lipsum}
\usepackage{enumitem}
\usepackage{caption}
\usepackage[margin=1in]{geometry}
\usepackage{marginnote}
\usepackage{dialogue}
\usepackage{hyperref}

\newcommand{\larry}[1]{{{\color{blue} #1}}}
\newif\ifinternal

\title{Why Build an Assistant in Minecraft?}
\author{Arthur Szlam,  Jonathan Gray,  Kavya Srinet, Yacine Jernite,  Armand Joulin,\\   Gabriel Synnaeve, Douwe Kiela, Haonan Yu, Zhuoyuan Chen, Siddharth Goyal, \\ Demi Guo, Danielle Rothermel, C. Lawrence Zitnick, Jason Weston}
\begin{document}
\pagenumbering{gobble}
\internalfalse

\maketitle

\begin{abstract}
In this document we describe a rationale for a research program aimed at building an open ``assistant'' in the game Minecraft,  in order to make progress on the problems of natural language understanding and learning from dialogue.
% In our view, the fundamental difficulty in such an endeavor is understanding what task a player might want the assistant to complete; whereas actually executing that task is relatively straightforward.  Thus, the central problem is of natural language understanding (NLU).  However, in service of the NLU problem, the game world can act as a shared resource between player and agent.
\end{abstract}

%The position being argued is that we should try to build an open assistant for the research community, and use it to make progress on the problems of task specification and learning from dialogue, without necessarily making progress on fundamental learning algorithms that allow one to learn everything end to end and scale up beyond an assistant in minecraft (We don't oppose working on those problems...).  We should use all the tools at our disposal to make task specification and learning from dialogue  as tractable as possible.  There is an argument to be made that we cannot make real progress in these without the giant end-to-end monster; but our hypothesis is its time to try. 

%how can we build behaviors and engineer components that allow flexible learning from humans (perhaps at the cost of scaling beyond a somehwat restricted scale) as opposed to "how can we make learning algorithms that can figure out how to improve themselves  and bounded only by data"

%"How can we use the environment as a resource for atttacking the NLU problem" (as opposed to "how can we demonstrate the effectiveness of end-to-end learning methods that can integrate various perceptual modalities"

\section{Introduction}\label{sec:introduction}
\enlargethispage{1cm}
In the last decade, we have seen a qualitative jump in the performance of machine learning (ML) methods directed at narrow, well-defined tasks. For example, there has been marked progress in object recognition \cite{mahajan2018exploring}, game-playing \cite{alphago}, and generative models of images \cite{karras2017progressive} and text \cite{jozefowicz2016exploring}.  Some of these methods have achieved superhuman performance within their domain \cite{alphago,atari}.   In each of these cases, a powerful ML model was trained using large amounts of %static 
data 
%or self-play  
on a highly complex task to surpass what was commonly believed possible.

Here we consider the transpose of this situation.  Instead of superhuman performance on a single difficult task, we are interested in competency across a large number of simpler tasks, specified (perhaps poorly) by humans.   In such a setting, understanding the content of a task can already be a challenge. 
%While in some situations a  specialized user interface can make this easier, and in others, the precision of a programming language is required, there are many situations where the most natural interface is through language.    
Beyond this, a large number of tasks means that many will have been seen only a few times, or even never, requiring sample efficiency and flexibility.% even if they are comprised of known components.  
%Consider the following dialogue between a human and their robot companion:
%\begin{dialogue}
%\speak{Human} hey bot lets play a game.  Every time I get close to you and say "cow" you have to jump twice.  if I say "pig" then you have to follow me, but really slow.
%\speak{Bot} ok
%\direct{plays game}
%\speak{Human} ok bot that was fun.  Lets call that game cowpig.
%\speak{Bot}  ok
%\direct{Bot can play game from name}
%\end{dialogue}

There has been measured progress in this setting as well, with the mainstreaming of virtual personal assistants.   These are able to accomplish thousands of tasks communicated via natural language, using multi-turn dialogue for clarifications or further specification.  The assistants are able to interact with other applications to get data or perform actions.  
%, and could be considered as agents embodied with an environment consisting of the   
Nevertheless, many difficult problems remain open.    Automatic natural language understanding (NLU) is still rigid and limited to constrained scenarios.      
Methods for using dialogue or other natural language for rich supervision remain primitive.  In addition,
because they need to be able to reliably and predictably solve many simple tasks, their multi-modal inputs, and the constraints of their maintenance and deployment,  assistants are modular systems, as opposed to monolithic ML models.  Modular ML systems that can improve themselves from data while keeping well defined interfaces are still not well studied.

%However, despite these advances in the potential components of a ``general'' AI system, there has been less success in combining these com.

%assistant, playmate, minion.

%we want to work in a setting where interaction is relatively easy

%In short: we are interested in the emergent properties of ML-augmented {\it systems}, and supervision from dialogue (beyond labels and reward). However,  despite the interesting dialogue-based ML-augmented systems in production,  opportunities for research using these systems is lacking.
%This will need to be reworded for external version, discussing the position of the paper esp the research program instead of the "garage" project

Despite the numerous important research directions related to virtual assistants, they themselves are not ideal platforms for the research community.  They have a broad scope and need a large amount of world knowledge, and they have complex codebases maintained by hundreds, if not thousands of engineers.   Furthermore, their proprietary nature and their commercial importance makes experimentation with them difficult.

In this work we argue for building an open interactive assistant, and through it, the tools and platform for researching grounded NLU. %dialogue-based ML-augmented systems.   
Instead of a ``real world'' assistant, we propose working in the sandbox construction game of Minecraft \footnote{Minecraft features: \textcopyright Mojang Synergies AB included courtesy of Mojang AB}.  
The constraints of the Minecraft world (e.g. coarse 3-$d$ voxel grid, simple physics) and the regularities in the head of the distribution of in-game tasks allow numerous hand-holds for  NLU research. Furthermore,  since we work in a game environment, players may enjoy interacting with the assistants as they are developed, yielding a rich resource for human-in-the-loop research.

The rest of this document describes in more depth the motivations and framing of this program, and should be read as a problem statement and a call-to-arms.  Concurrently, we are releasing \url{https://github.com/facebookresearch/craftassist}, which houses data and code for a baseline Minecraft assistant, labeling tools, and infrastructure for connecting players to bots; we hope these will be useful to researchers interested in this program or interactive agents more generally.  We detail the contents of the framework (\url{https://github.com/facebookresearch/craftassist}) in \cite{FrameworkPaper}.

\section{Minecraft}
Minecraft\footnote{\url{https://minecraft.net/en-us/}} is a popular multiplayer open world voxel-based building and crafting game.
%\larry{Add screen captures from gameplay.} 
Gameplay starts with a procedurally created world containing trees, mountains, fields, and so on, all created from an atomic set of a few hundred possible atomic blocks; and (atomic) animals and other non-player characters (collectively referred to as ``mobs''). 
%\larry{Add figure showing block types.}
The blocks are placed on a 3D voxel grid. Each voxel in the grid contains one material, of which most are air (empty); mobs and players have floating point positions. 
Players can move,  place or remove blocks of different types, and attack or be attacked by mobs or other players. 
%e.g., a house could be constructed from brick, glass and door blocks. 

The game has two main modes\footnote{There are 5 modes in total: survival, hardcore, adventure, creative, and spectator.}: ``creative'' and ``survival''. In survival mode the player is resource limited, can be harmed, and is subject to more restrictive physics. The player must gather and combine resources in a process called “crafting” to build objects. In creative mode the player is not resource limited, cannot be harmed, and is subject to less restrictive physics, e.g., the player can fly through the air. Crafting is not necessary in creative mode since all materials are available to the player. An in-depth guide to Minecraft can be found at \url{https://minecraft.gamepedia.com/Minecraft}.
%In this paper, we focus on interactive agents in creative mode. However, the same infrastructure could also be used in survival mode.
Building is a core gameplay component in both modes.

Minecraft allows multiplayer servers, and players can collaborate to build and survive, or compete. It has a huge player base (91M monthly active users in October 2018) \cite{MCMAU}, and many players are active in creating game mods and shareable content.  The multiplayer game has built in text chat for player to player communication. Dialogue between users on multi-user servers is a standard part of the game.

%Players are encouraged to be creative. The diversity of objects created in Minecraft is astounding; these include landmarks, sculptures, temples, rollercoasters and entire cityscapes. Collaborative building is a common activity in Minecraft.

\subsection{Task Space} %Minecraft Assistant task space}
\label{sec:task_space}
Minecraft players can build things, gather resources, craft (combine resources), attack other players and mobs (non-player characters), and chat. Even focusing only on building, the set of things a player could possibly do in the game is enormous; in the most naive sense, it is all possible ways of placing all the possible blocks into as big a world as fits in RAM.  Minecraft players are creative, and the diversity of player built objects in Minecraft is astounding; these include landmarks, sculptures, temples, rollercoasters and entire cityscapes. Collaborative building is a common activity in Minecraft.

%{\color{red} Screenshots, examples here?}

   Nevertheless, the Minecraft task space is constrained, due to restrictions on the environment and player behavior.    Physics in Minecraft is particularly simple, and as mentioned above, the world is arranged on a coarse 3D grid.  There is a finite (and relatively small) list of atomic game objects (mob types, block types, etc), and a small finite list of crafting formulae.
%A Minecraft assistant could potentially assist in any of these tasks by performing sequences of relevant actions instead of the player performing them, given a clear and succinct set of natural language commands issued by the player via chat.
We expect that
 the distribution of player requests of an assistant will  be concentrated on a tiny fraction of what is actually possible in the game.   For example, the vast majority of block arrangements are unlikely building requests, and modern ML has shown some success at being able to learn to sample (or retrieve) perceptually pleasing 3d structures \cite{sung2017complementme}.     Because an assistant bot could already be helpful by successfully completing common tasks from the head of the distribution even if it fails  on tasks from the tail, we believe we can make progress towards a useful assistant without having to be able to succeed at every possible request. 
If true, this could pave the way to further learning during deployment.%,as interaction with human players will already take place.

The constraints of the environment allow that {\it executing} a task is straightforward once it is specified.  Movement, construction, crafting, and even scouting and combat can be reasonably scripted.  While there might be value in learned representations (of language, the environment, actions, etc.) that come from learning to execute, it is not {\it necessary} to use ML to solve task execution.   On the other hand, because it is possible to script execution, to the extent that it useful to get learned representations, we can easily get supervision to train these.

\begin{comment}
\subsection{Action primitives}
Describe the set of actions a Minecraft player can perform. Add more details below.
\begin{enumerate}
\item Move.
\item Place a block.
\item Destroy a block.
\item Attack a mob
\item Gather or pick up a resource.
\item Point (only available to the bot?)
\item Eat?
\item Sleep?
\item Craft
\item Talk
\end{enumerate}

\subsection{Player inputs}
Explain what a player or bot could observe: perception and dialog.

\subsection{Player tasks}
Describe common tasks performed by players (resource gathering, house building, crop growing, etc.).

%Build things
%Craft things (collect and transform resources)
%55M monthly active players in February 2017 (edit: 91M in October 2018)
%Multiplayer servers with chat
%Lots of mods and user created content
%Used to teach kids to program! (e.g. https://minecraft.makecode.com/)
\end{comment}

\begin{comment}
\subsection{Why Minecraft for studying intelligent assistants?}
The open world setup of Minecraft is ideal for studying assistants that can solve a large variety of nontrivial tasks.   We want to emphasize collaboration with humans, and gameplay centered around building and less focused on violence is attractive.  The large player base make it possible to have bots interacting with people at scale.
\end{comment}

% LocalWords:  Minecraft multiplayer voxel Gameplay procedurally NPC
% LocalWords:  storyline rollercoasters cityscapes shareable multi
% LocalWords:  NPC's Ender minigames spleef gameplay

\newcommand\botname{assistant}
%\widowpenalty10000
\section{Learning and Interaction in Minecraft}

We wish to build an intelligent in-game assistant\footnote{or playmate, or minion, etc...} that can perform whatever players may want it to do (in-game). 
% We want to take advantage of the successes of modern ML, but also, we  intend to use as much domain knowledge as is necessary.
%The assistant will be instantiated as another player in the game (and so have all the in-game capabilities of the player), but
% be controlled by our system.   
 Simple examples of the assistant's duties could range from building simple or complex structures to entire cityscapes,
breaking down structures, dancing, catching mobs (Minecraft creatures), etc.
We intend that the assistant's primary interface should be via natural language using Minecraft chat.
Finally, we aim for an agent that is fun, and one that people will want to play and engage with.  The purpose of building such an agent is to facilitate the study of the following:
\begin{itemize}
\item
\noindent {\bf Synergies of ML components:} To explore and evaluate various approaches to building a complex agent. In particular, how the various ML and non-ML components of a system can work together, and how to exploit the synergies of these components.  Especially, how to exploit these components to make progress in:
%\newline
\item
\noindent {\bf Grounded natural language understanding:}
Here, specifying tasks is the challenge, not executing them.  How can we build a system to understand what a human wants; and to associate language to a task?   
%advance the state of the art in grounded natural language understanding; and to provide a platform for ourselves and studying  grounded natural language understanding, together with our learning agents and training data, in order for the community to make fundamental advances.
\item
%\newline
\noindent {\bf Self improvement:}
To make progress in building agents with the ability to improve themselves during deployment and interaction with human players.  How to learn new tasks and concepts from dialogue or demonstrations?
\end{itemize}

The basic philosophy of this program is that we should approach the relevant NLU and self-improvement problems by all means available.   In contrast to
 many agent-in-an-environment settings, where the environment is a challenge used to test learning algorithms, we consider the environment a tool for making the NLU problems more tractable.  Thus instead of ``what ML methods can learn representations of the environment that allow an agent to act effectively?'' we are interested in the problem of ``what approaches allow an agent to understand player intent and improve itself via interaction, given the most favorable representations (ML based or otherwise) of the environment we can engineer?''.   While we are sympathetic to arguments suggesting that we will be unable to effectively attack the NLU problems without fundamental advances in methods for representation learning\footnote{And we think programs aimed at advancing these are great too!}, we think it is time to try anyway.
 
Code and other tools to support this program are available at \url{https://github.com/facebookresearch/craftassist}.
 
\subsection{Natural Language Interaction}

The assistant should interact with players through natural language.
This is not meant to handicap the assistant, and other methods of interaction may also be helpful.  However,  we believe the flexibility and generality of  language will be useful (and perhaps necessary).  Natural language allows that players will need no specialized training to play with the assistant.  Furthermore, combined with the fact that the agent is just another player in the game, this makes training and evaluation scenarios where a human pretends to be a bot straightforward.

We intend that the player will be able to specify tasks through dialogue (rather than by just issuing commands), so that the agent can ask for missing information, or the player can interrupt the agent's actions to clarify.    In addition, we hope dialogue to be useful for providing rich supervision.   The player might label attributes about the environment, for example ``that house is too big'', relations between objects in the environment (or other concepts the bot understands), for example ``the window is in the middle of the wall'', or rules about such relations or attributes.  We expect the player to be able to question the agent's ``mental state'' to  give appropriate feedback, and we expect the bot to ask for confirmation and use active learning strategies.

We consider the natural language understanding (NLU) problem to be the program's central technical challenge.  We give some examples in the next section \ref{sec:nlp_challenges}.
While a full solution is likely beyond the reach of current techniques, working with an interactive agent in Minecraft allows opportunities for progress.
We can rely on agent grounding (a shared knowledge resource between player and agent, in this case the semantics of Minecraft) to aid learning language semantics. Moreover, we expect to spend lots of human effort on making sure common requests are understood by the bot, in order to bootstrap interaction with players.  We discuss these in more detail in section \ref{sec:nlp_opportunities}

\subsubsection{Challenges}
\label{sec:nlp_challenges}
The set of things that Minecraft players do for fun (building, crafting, resource gathering) is combinatorially complex.
 Even focusing on creative mode (where crafting and resource gathering are not relevant dynamics), there is a huge space of things a player could ask an assistant for help with.  We give some examples and discuss their complexities:
 \begin{dialogue}
\speak{player} build a tower 15 blocks tall and then put a giant smiley on top
\speak{\botname}  ok
\direct{\botname\ starts building the tower}
%\speak{player} wait, stop, make every other block red
\end{dialogue}
To succeed in executing this command, the assistant needs to understand what a ``tower'' is (and how to build one),  understand that ``15 blocks high''  measures  the height of the tower, and what ``15'' is.   It needs to know what a ``smiley'' is (and how to build it) and understand the relative position ``top''.     In our view,  with the assumptions of this program, this is nontrivial but straightforward, for example using techniques as in \cite{Zelle:1996, ZettlemoyerC05,P11-1060, Liang:2016}.  %explain why?
However, it is easy to imagine small changes that make it more difficult:
 \begin{dialogue}
\speak{player} build a tower 15 blocks tall and then put a giant smiley on top
\speak{\botname} ok
%\botname}  ok
\direct{\botname\ starts building the tower}
\speak{player} wait, stop, make every other block red
\end{dialogue}
Besides needing to know what ``every other'' means, in this scenario, the assistant has to recognize that the player is referencing a change to the tower and smiley that it is currently building, and that ``stop'' doesn't mean ``stop activity''.  The assistant also needs to know what ``red'' is but that is more straightforward.  If the assistant does not know what ``every other'' means, we could imagine a dialogue like
 \begin{dialogue}
\speak{player} build a tower 15 blocks tall and then put a giant smiley on top
\speak{\botname}  ok
\direct{\botname\ starts building the tower}
\speak{player} wait, stop, make every other block red
\direct{\botname\ recognizes the instruction refers to a change in its current build task, but doesn't understand the change (and specifically recognizes ``every other'' as unknown)}
\speak{\botname} What is ``every other''?
\speak{player}  Let me show you
\direct{player makes a stack of blocks alternating in color; assistant is able to generalize ``every other'' to new situations}
\end{dialogue}

In our view, this level of flexibility and ability to learn from dialogue and examples is beyond what we can accomplish with our current technology. %and the scope of the project.
However, we believe that situating the assistant in an engaging environment that allows simplified perception and task execution opens opportunities to
%  first reach the point where we have built an assistant framework and a level of user engagement where we can 
do research at this frontier.

\subsubsection{Opportunities}
\label{sec:nlp_opportunities}
As mentioned in \ref{sec:task_space}, the Minecraft task space and environment have many regularities that we can use to simplify task execution.  These regularities also create handholds for the NLU problems of specifying tasks and learning from dialogue.   
 
 %\larry{We should be more explicit about what areas in NLP will most benefit, i.e., grounding, mapping language to action, VQA, etc.}: the goal is specifically to make progress on the bullets form the above: task specification and learning from dialogue

 First, they allow us to engineer language primitives or otherwise introduce domain knowledge into data collection and model design.  For example, using the knowledge of the basic Minecraft tasks, we can build sets of language/action templates for generating examples commands for these tasks (in addition to crowdsourcing such language).   %In particular, the action primitives in \ref{sec:action_primitives} are amenable to forming a simple grammar which we use in our V0 bot, and is described in \ref{sec:action_dictionaries}.
 These can be used to build artificial training data and inform  the structure of machine learning models that are meant to interpret language.  Another example is basic behavioral patterns that elicit data from the user, like asking a player to tag things it sees in the environment, which can be later used as referents.   %We want to emphasize that these are particular examples we have already employed, but 
 There is a huge space of opportunities for endowing our assistant with a range of language primitives.  These allow building useful assistants from which to bootstrap before being able to learn too much from players.

Second, the structure of the environment and the head of the task distribution will hopefully allow the assistant to learn language more easily, beyond data generation and model design.   This structure can function as a knowledge resource shared between agent and player, and ground concepts the assistant might need to learn.    For example, if the user asks the assistant to ``build a smiley'', the agent can infer that ``a smiley'' is some kind of block object, as ``build'' is a common task the bot should already understand.  The agent can make connections between ``small'' and the number of blocks in an object, and ``close'' and the number of steps it needed to walk.  Also, atomic Minecraft objects give a set of reference objects with rich structure that require no learning to apprehend: mob types, block types etc. are given explicitly to the assistant.
Furthermore, we might expect to be able to find synergies between learning generative models of block objects and discriminative models of those block objects (and their parts) in addition to the language used to intruct the assistant to build or destroy those objects.
In our view, one of the most exciting aspects of building an assistant in Minecraft is this grounding of concepts afforded by the environment and the distribution of tasks. %our conjectures on the distribution of tasks.

Third, the basic game is {\it fun}, and gives us opportunities to understand how to make training (or teaching) an assistant rewarding.  We expand on this more in \ref{sec:fun}.

\subsection{Self improvement}
\enlargethispage{1cm}
One of the oft repeated criticisms of standard statistical ML is that models cannot improve themselves beyond becoming more accurate at the task for which they were designed (and for which the data has been collected and labeled, etc.).  They cannot reconfigure themselves for new tasks without a human building a new dataset (or reward function).   While this complaint is not very precise, we do believe it is true in spirit, especially for the most successful ML  models, such as those used in object recognition, translation, or language modeling.   While the features they produce can be useful in many tasks, the model themselves are inflexible.

On the other hand, there are several systems designed explicitly for self improvement, for example \cite{NELL-aaai15, SidaNaturalizing}.  We consider the setting of a Minecraft assistant an ideal setting for studying how to flexibly learn from human interaction.  As in \cite{NELL-aaai15, SidaNaturalizing}, instead of trying to ``solve'' self-improvement by making fundamental advances in ML, we think the setting will allow progress because it allows the correct substrate.  That is, we hope we can design frameworks allowing an assistant to improve itself within that framework (but not beyond that);  and because of the richness of the Minecraft task space, this allows for non-trivial growth.

\subsubsection{Effective (and fun)  training}
\label{sec:fun}
%Minecraft is a good place to study how to make training fun
One of the goals of the program is to study how the assistant should interact with people, and what feedback mechanisms should be employed to make training the assistant both efficient and enjoyable.  While efficient learning is probably the key technical ML challenge to making the assistant responsive to training, there are many other UI factors that may be as important for making training fun.  As part of the project, we want to study how the assistant should respond to feedback so that it is engaging.  Moreover, mistakes should be amusing, and the assistant should more generally fail gracefully.  These criteria are important in order to incentivize players to both engage with the assistant and to enable it to receive a learning signal to improve itself.  

In terms of efficiency (in ML terms, being very sample efficient) and user enjoyment, some major challenges include:
\begin{itemize}
\item For the assistant to be immediately responsive to feedback. We suspect this will have a large effect on the players impression of the bot's performance.
\item To optimally interact with the user, for example seeking clarification that both gives maximal information and does not bother the player with a succession of annoying questions.
\item For the assistant to get the most out of the (rich) feedback that is given \cite{Srivastava2017,BradenBabble}, i.e. to generalize maximally beyond the individual case being addressed.
\end{itemize}
%A second major challenge, is for the assistant to get the most out of the (rich) feedback that is given \cite{Srivastava2017,BradenBabble}

 So, for example, we can imagine the following dialogue:
 \begin{dialogue}
\speak{player} go over to that tower with the gargoyles
\direct{\botname\ does not know what ``gargoyles are'', but sees there is only one tower; surmise that that tower must be related to ``gargoyles''}
\direct{\botname\ walks to the tower}
\speak{\botname}  does this tower have gargoyles?
\speak{player} yes
\speak{\botname}  What is a gargoyle?
\speak{player} its the monster on top of the tower
\direct{\botname\ recognizes is it is a part of the tower because of ``top'' which it already understands}
\speak{\botname}  is it this?
\direct{\botname\ moves to the top of the tower}
\speak{player} yes
\end{dialogue}
The information the assistant could gain from this dialogue could be learned by seeing lots of objects labeled with ``gargoyle'', or associating a reward given by a player when it moved to the correct object over lots of plays.  But in our view, if the assistant could execute these sorts of language interactions, it could be far more sample efficient.  In this case, through the interaction, the agent could learn that a gargoyle is an object made of blocks (and not some other property of a tower, like being tall), find an example of this kind of object and a rough segmentation, and an association between ``gargoyle'' and the word ``monster''.
%Even if it could not generalize ``gargoyle'' to a new instance, it could memorize that instance and place it if the user asked it to build another tower with a gargoyle on it.

As before, this level of dialogue is still science fiction, and is for illustrating the opportunities of rich feedback.  However, one can see that in order to even study these kinds of interactions in a non-toy setting with our current technology, we need an environment where the assistant has a good deal of domain knowledge built in.  We also need players who are motivated to teach the assistant.   We think it will be useful (and interesting) to design learning behaviors explicitly in addition to trying to learn them.
\begin{comment}
\subsubsection{Pointing and other UI elements}
In the above dialogue, it would have been useful for the assistant and the player to be able to point.  It turns out that this is not difficult in Minecraft.   For user pointing, we can use the position of the user's crosshairs when they enter a chat.   For the assistant pointing, we can rapidly place and remove blocks, resulting in a flashing effect.   To the extent it is necessary or useful, we consider it in the project's scope to investigate other UI elements for ``non-verbal'' communication in game in the future. \larry{We should add this to the action dictionary section.}
\end{comment}
%{\color{red} talk about pointing, modifying blocks, other ``non-verbal'' communication}

\subsection{Modularity}
%\enlargethispage{1cm}
Many of the recent successes of ML have come from training end-to-end models.    Nevertheless, we believe that in this research program, the best path forward is to 
 bootstrap a modular system with both scripted and ML components.    We are not {\it opposed} to end-to-end approaches, and think it may be possible for an ML agent to learn to do everything end-to-end from raw pixel input.   However, our view is that this too difficult in the near term,  %unecessary 
       and that the challenges of learning low-level actions and perception are likely to be a distraction from the core NLU and self-improvement problems.   In contrast, a modular system allows us to abstract away the lower-level primitives and focus on these problems.  Furthermore, using a modular system makes data collection and generation easier.   Finally, besides making this research program more accessible, we think modular ML-augmented systems are generally interesting objects of study.  
 
 In particular, we do not consider it ``cheating'' to script useful action, perception, or symbolic manipulation primitives.   These might include path-finding and building scripts, and  libraries of schematics and shapes; or heuristics for relative directions or sizes of objects.   Similarly, we consider it reasonable  to use ML components trained for a particular sub-task (perceptual or otherwise).   Our frameworks (as in \cite{gehring2018torchcraftai}) and models (as in \cite{andreas2016neural, gehring2018torchcraftai, lee2018modular}) should be designed to allow for a gradual process of ML-ification where it may be useful.  
 
\medskip  
        
\noindent {\bf Abstracting lower-level primitives:}
The fundamental problem that a Minecraft assistant faces is understanding what the player wants the assistant to do.
In contrast, for many requests, execution is more straightforward.
In Minecraft, the sequence of (atomic) actions necessary to complete basic tasks (path-planning, building, etc.) can be scripted by directly accessing the game's internal world state.
 This is clearly not true more generally -- in the real world, or even in the settings of other games, the execution of similar basic actions cannot be easily scripted.   
 %For example, in Minecraft, %placing a block at a given location is a single atomic action; whereas having a real robot learn to place a physical block is still not a solved problem.  Path-finding in Minecraft 
% building a known schematic can be easily scripted.
 %; but could be quite complicated in other settings.   
  Furthermore, many of the easily scriptable action primitives in Minecraft (like building) might take hundreds or even thousands of steps to complete, making them non-trivial to learn.    Similarly, many perceptual primitives are straightforward to implement, despite being non-trivial to learn.  For example, mobs are atomic objects, as are blocks, despite having surface forms (as images) that vary based on viewing angle, occlusions, lighting conditions, etc.
 In this research program, we consider these and other particulars of the environment as tools for directly engaging the fundamental problem of player intent. 
  
\medskip  
  
\noindent {\bf Gathering data:}
The successes of end-to-end models have been driven by large data; but in this program, there is no (initial) source of data for end-to-end training.   The end-to-end task requires human interaction, and with our current technology, agents are not responsive enough to supervision to learn in a way that would be engaging.  Again because of the human in the loop, self-play is not straightforward.    Simply recording human players playing normally (without an in-game assistant) for behavioral cloning  is not ideal, as standard play is different than what we would want the assistant to learn; and having humans play as assistants with other humans does not easily scale.   
%   The domain knowledge necessary to build the environment setup and the reward function(s) for e.g.  reinforcement learning for sub-tasks is often no more than what is required to the sub-tasks easy.    
On the other hand, for many action or perception primitives, it is easy to collect or generate data specifically for that primitive.
  %we consider this an unnecessary
These considerations suggest approaches with modular components with clearly defined interfaces,  and heterogeneous training based on what data is available (in particular, we are interested in models like \cite{andreas2016neural} that can be trained end-to-end as a whole or separately as components).
  
\medskip  
  
%Modular, ML-augmented systems are standard in commercial deployment, but  .    
\noindent {\bf Modularity as a more generally useful ML trait:}
We consider the study of ML-augmented systems as an interesting endeavor in its own right.   
An open assistant can be used as a laboratory for studying the interactions between ML and non-ML components and how changing one component affects the others.  Or how to design and build systems so that their behavior can be easily modified, explained, and engineered,  while still having the capability to flexibly learn.  While the specific primitives for this program should be designed with the Minecraft assistant in mind,  we expect that many of the things we discover about building systems will be useful more generally.

  \medskip
  
Some researchers might argue that without the fundamental learning algorithms that would allow an agent to build its understanding from the ground up, we cannot scale beyond simple scripts and hand-designed behaviors.   We agree that this is a risk, and we certainly agree that scripts and hard-coded perceptual primitives will not be sufficient for real progress\footnote{However: while there can be important generalization and flexibility benefits from an end-to-end system, so too can there be different generalization and flexibility benefits to having the correct heuristic/symbolic substrate.}.   Our view is that the NLU and self-improvement problems  have difficulties beyond representation learning that should be confronted as directly as possible, and scripted or separately trained components should be used to the extent they are useful in making progress in these core problems.

\section{Literature Review}

\noindent {\bf Existing Research Using Minecraft}
A number of machine learning research projects have been initiated in Minecraft.
Microsoft has built the MALMO project \cite{johnson2016malmo} as a platform for AI research.
It is often used as a testbed for ML model architectures trained using reinforcement learning methods, e.g.
 \cite{shu2017hierarchical,udagawa2016fighting,alaniz2018deep,oh2016control,tessler2017deep}.  Some of these use (templated) language to describe mid-level 
 macros \cite{shu2017hierarchical} or tasks \cite{oh2017zero}.
The work of \cite{lin2017explore} considers the use of human feedback, but at the action level, not by via language.
The Malmo Collaborative AI Challenge\footnote{\url{https://www.microsoft.com/en-us/research/academic-program/collaborative-ai-challenge/}} was recently proposed to explore the training of collaborative RL agents, but does not address collaboration with humans, nor the use of language.  
Most recently \cite{MineRL} collects millions of frames of players doing various tasks in Minecraft, and proposes several (single-player) tasks to evaluate RL and imitation learning methods.
\cite{yi2018neural} considers the use of language in Minecraft to answer templated  visual questions.
\cite{kitaev2017misty} focuses on a dataset and neural model for spatial descriptors (``on top of'', ``to the left of'').

An observational study of how humans would speak to an intelligent game character in Minecraft using a Wizard of Oz approach has been recently conducted \cite{allison2018players}.  
%In summary, existing work using Minecraft  (1) does not consider the use of natural language; and (2) does not consider learning to collaborate with human partners, e.g. as an assistant. They thus do not take advantage of the fact that Minecraft has a large player base of humans combined withe ability to chat with them, two ingredients we see as special about the platform.
%Note that 

\medskip  
  
\noindent {\bf Other Gaming Platforms and Simulated Environments}
A number of other games are used as platforms for AI research.
Many are built to study the development of reinforcement learning based algorithms and do not study language, e.g. Starcraft \cite{synnaeve2016torchcraft,vinyals2017starcraft},
Atari games \cite{bellemare2013arcade}, Go \cite{silver2016mastering,tian2017elf},
 TORCS \cite{wymann2000torcs}, Doom \cite{kempka2016vizdoom} and Text adventure games \cite{cote2018textworld,yang2017mastering}.
DeepMind Lab \cite{beattie2016deepmind} is based on the Quake engine, although it is somewhat removed from the game itself.
%Some of those games have resulted in research  focused on  learning from raw pixel inputs and fine motor dexterity (TORCS, Atari, DeepMind Lab, VizDoom), while the first-person setups additionally focus on navigation. This is in contrast to our proposed research program which ignores those directions in favor of grounding actions with human collaboration and communication.
Some of these systems do not involve a human at all, e.g. one-player Atari Games, while others such
as Starcraft and Go do involve an agent interacting with a human, but in an adversarial fashion, rather than as an assistant. 
%Nevertheless, they are often trained in self-play mode or on historical human-human games rather than during their interactions with humans.

Instead of directly using an existing game platform, another approach has been to implement a simulation for embodied agent research, especially for navigation, QA and situated dialogue.
Several such environments have been proposed, such as 3D environments
like  House3D \cite{wu2018building}, HoME \cite{brodeur2017home}, MINOS \cite{savva2017minos}, Matterport3D \cite{chang2017matterport3d}, AI2-THOR \cite{kolve2017ai2}, and Habitat \cite{habitat19arxiv}, 
as well as more simplistic 2D grid worlds \cite{sukhbaatar2015mazebase,yu2018interactive, chevalier2018babyai}.
Within those environments typical tasks to study are language grounding, navigation,
embodied QA\cite{das2018embodied}.  The visual content of Minecraft is much less realistic than \cite{wu2018building,brodeur2017home,savva2017minos, chang2017matterport3d, kolve2017ai2}; but the task space is richer.
In \cite{bisk2016natural,  bisk2018learning}, the authors explore natural language instruction for block placement, first in 2D and then in 3D.    Similar to the setting we consider, visual fidelity is secondary to natural language complexity, and their environment is focused on building instruction, although there the agent is not able to navigate or dialogue with players.    One crucial difference between all of these environments and Minecraft is that Minecraft is a engaging game that already has a huge player base, hopefully enabling the study of learning agents that interact with humans via natural language at scale.  

To a large extent, all the above platforms have been used mostly to answer questions of the form ``how can we design algorithms
and architectures that are able to learn via acting in an environment (perhaps integrating multiple modalities)?''
In particular end-to-end learning and reinforcement is emphasized, and ``cheating'' by explicitly using domain knowledge
or directly using the underlying mechanics of the simulator is discouraged.  The agent is supposed to learn these.
In our program,  we wish to answer questions of the form "how can we use the simulator as a knowledge resource, shared between
player and learning agent, in service of understanding intent from language?".  Anything that makes this easier is fair game, including
directly using the underlying game objects and domain knowledge about the environment and task space.

%The approach advocated here can similarly be contrasted with \cite{mikolov2016roadmap}, which 

%%%%%TODO for external
\begin{comment}
\paragraph{Robotics}
Fill in.
Bwibots platform designed for AI and human–robot interaction research \cite{khandelwal2017bwibots}.
In contrast Minecraft is a simulator with massive human engagement, making it much
simpler and cheaper platform from which to conduct research.
Nevertheless, our hope is that some of the learnings from building Minecraft assistants will  generalize to systems in the real world.
\end{comment}

\medskip  
  
\noindent {\bf Personal Assistants and dialogue}
Virtual personal assistants have now penetrated the consumer market, with products such as Google Assistant, Siri and Alexa. These systems are grounded in world knowledge; and can be considered ``embodied'' in an environment consisting of web and device responses.  They could  be seen as an important platforms for research, especially in terms of dialogue and human interaction.
Unfortunately, as they are large proprietary production systems, and as experimentation can negatively affect users and brands,  they are not easily used in
 open (academic) AI research.
Currently, such academic research typically involves smaller, less open-ended domains.
There is a large body of work on goal-oriented dialogue agents
which typically focus on one task at a time, for example
restaurant booking \cite{williams2013dialog,henderson2014second,bordes2016learning},
 airline booking \cite{asri2017frames,wei2018airdialogue}, or movie recommendation \cite{li2018towards}.
Minecraft as a platform is relatively a good setting for goal directed dialogue research, as
(1) a Minecraft agent is embodied in an environment with many familiar physical and perceptual primitives, and so offers opportunities for correlating language, perception,  and physical actions; and
(2) due to its open-ended nature, the setting emphasizes competency at a wide variety of tasks; 
(3) the game itself is engaging, and
(4) amusing enough mistakes are more tolerable.

\medskip  
  
\noindent {\bf Semantic Parsing and Program Synthesis}
Many of the simpler goal-oriented dialogue tasks discussed above can be formulated as slot filling, which can be considered a form of semantic parsing (inferring a machine understandable logical form from a natural language utterance).  Semantic parsing has been used  for interpreting natural language commands for robots \cite{tellex2011understanding, matuszek2013learning} and for virtual assistants \cite{kollar2018alexa}.  
 Semantic parsing in turn can be considered a form of the general problem of program synthesis (see \cite{gulwani2017program} for a survey), where the input to the program synthesizer is a natural language description of what the program should do.   Recently there have been many works applying ML to semantic parsing \cite{artzi2013weakly, liang2016neural, dong2016language, jia2016data, guu2017language, zhong2017seq2sql} and program synthesis from input-output pairs \cite{lake2015human, reed2015neural,gaunt2016terpret,neelakantan2016learning, bovsnjak2017programming,gaunt2017differentiable} (and non-ML success stories \cite{gulwani2012spreadsheet}).

An assistant in Minecraft offers the opportunity for studying program synthesis through interactive dialogue and demonstrations; and in a setting where the agent can learn over time about a player's particular task distribution and language.   These have been less studied, although see \cite{iyyer2017search, guo2018dialog}  for dialogue based semantic parsing and \cite{mayer2015user} for multi-turn program synthesis and  \cite{ gaunt2016lifelong, wang2017naturalizing} for ``lifelong'' learning to synthesize programs.

\medskip  
  
\noindent {\bf Learning to Ground Language outside of Simulators}
There are numerous works exploring the link between modalities in order to ground language outside of a simulator by considering static resources, e.g. by combining vision and text.
Such works typically consider a fixed database of
images and associated text, e.g. for
image captioning \cite{lin2014microsoft},
 video captioning \cite{yu2016video}, visual QA \cite{antol2015vqa} or
visual dialogue \cite{das2017visual,shuster2018engaging}.

\medskip  
  
\noindent {\bf Learning by Instruction and Interactive Learning}% and Never-Ending Learning}
We are interested in agents that learn from feedback from the user.
%There is a body of work for learning language during deployment for unembodied dialogue agents. %:  chatbots and goal-oriented dialogue agents.
 Learning from dialogue with varying types of feedback, such as verbal cues (e.g., ``Yes, that's right!'') was explored for the question answering (QA) setting in \cite{weston2016dialog} and \cite{li2016dialogue}, and in a chit-chat setting in \cite{hancock2019learning}.
 %using novel learning approaches such as a form of forward prediction for learning.
Similar techniques have been applied for teaching machines to describe images via natural language feedback as well \cite{ling2017teaching}.
Agents can gather more information by also proactively asking questions rather than just waiting for feedback. This has been studied in (ungrounded) conversational systems \cite{strub2017end,wang2018learning,rao2018learning,li2016learning} and when grounding with images via the CLEVR QA task \cite{misra2017learning}.
Several works have studied how to use language to train a parametric classifier, e.g. \cite{srivastava2018zero, BradenBabble, elhoseiny2013write}.
The topic of learning by instruction was also the subject of a NIPS 2018 workshop\footnote{\url{https://nips.cc/Conferences/2018/Schedule?showEvent=10918}}.
While there is some work in learning from dialogue in an embodied setting \cite{cantrell2011learning,thomason2017guiding}
this is currently relatively unexplored,
%To our knowledge, there is so far less work in learning such feedback in am embodied setting,
%primarily because that requires humans-in-the-loop which many platforms cannot provide. %Minecraft appears to be a good setting for this line of research.
%
The program described in this work is related to (and has been inspired by) \cite{wang2016learning,wang2017naturalizing}. In \cite{wang2016learning} an interactive language learning game is set up between human and machine in order to place blocks appropriately given a final plan. In \cite{wang2017naturalizing} a more open ended building task is proposed using a grid of voxels.  The machine has to learn the language used by humans in a collaborative setting to perform well specified but complex actions in order to draw shapes out of blocks.

We have also been inspired by \cite{NELL-aaai15}, \cite{mikolov2016roadmap}, and \cite{lake2017building}
and in particular, we share the goal of exploring agents that can learn autonomously from various forms of supervision beyond labels.  Perhaps in a naive sense, what we are suggesting is less ambitious than these:  instead of learning about the real world as in \cite{NELL-aaai15}, the goal is to learn how to do tasks in Minecraft (and process the relevant concepts) that are likely to be given to Minecraft assistant.  Instead of working towards a method that can learn in an unbounded way, as in \cite{mikolov2016roadmap}, we aim ``only'' for learning within the Minecraft frame, with as advantageous a perceptual and knowledge-representation substrate as we can find.
% However, we are less focused on long term knowledge accumulation building world models, and instead of learning about the real world through interaction with the web, the goal is to learn about the Minecraft world through interactions with players.    
The focus in this work on  task specification and learning through language in a known environment recalls the ``Frostbite Challenge'' in \cite{lake2017building}; but we do not consider it important whether our agents have human biases.    Nevertheless, it may turn out that succeeding in the Minecraft task space requires understanding the real world, that in order to be able to learn as flexibly as we would like in the Minecraft frame necessitates methods that can learn as desired by \cite{mikolov2016roadmap}, and that in order to successfully complete human tasks and learn from human instruction, the agent will need human biases.   In any case, we think the setting of an interactive assistant in Minecraft is an ideal frontier to work on these problems.

% LocalWords:  Minecraft cityscapes lookups generalizable Screenshots
% LocalWords:  useable NLP combinatorially ok crowdsourcing SOTA DL
% LocalWords:  SHRDLU incentivize bot's UI crosshairs dataset Siri RL
% LocalWords:  personalizations gameplay featurize NPCs orientable QA
% LocalWords:  modalities modality prebuilt behaviour MALMO templated
% LocalWords:  Malmo Starcraft TORCS DeepMind formulae VizDoom HoME
% LocalWords:  Matterport Bwibots learnings Alexa orientded CLEVR
% LocalWords:  unembodied ungrounded voxels

\section{Conclusion}
In this work we have argued for building a virtual assistant situated in the game of Minecraft , in order to study learning from interaction, and especially learning from language interaction.  We have argued that the regularities of the game environment and task space can and should be used explicitly as tools to make the relevant learning problems more tractable, rather than as diagnostics to measure if a method can learn those regularities.   

We hope the broader research community will be interested in in working on this program, and to to that end we open \url{https://github.com/facebookresearch/craftassist} with infrastructure for connecting bots to players, and data and code for baseline bots.

%Automatic natural language understanding remains inferior to human natural language understanding, even in relatively constrained environments, and methods to use the full richness of language supervision remain primitive.  We propose a program of building an assistant in Minecraft

\bibliography{refs}
\bibliographystyle{plain}

\end{document}